\documentclass[sigconf,dvipsnames,nonacm]{aamas} 

\usepackage{balance} 

\usepackage{bm}
\usepackage{algpseudocode}
\usepackage{amsmath}
\usepackage{graphicx}
\usepackage{lipsum}
\usepackage{subcaption}

\newcommand{\mc}[1]{\mathcal{#1}}

\setcopyright{ifaamas}
\acmConference[AAMAS '24]{Proc.\@ of the 23rd International Conference
on Autonomous Agents and Multiagent Systems (AAMAS 2024)}{May 6 -- 10, 2024}
{Auckland, New Zealand}{N.~Alechina, V.~Dignum, M.~Dastani, J.S.~Sichman (eds.)}
\copyrightyear{2024}
\acmYear{2024}
\acmDOI{}
\acmPrice{}
\acmISBN{}

\acmSubmissionID{814}

\title[AAMAS-2023 Formatting Instructions]{Deceptive Path Planning via Reinforcement Learning \\with Graph Neural Networks}
\author{Michael Y. Fatemi}
\affiliation{
  \institution{University of Virginia}
  \city{Charlottesville, VA}
  \country{United States}}
\email{gsk6me@virginia.edu}

\author{Wesley A. Suttle}
\affiliation{
  \institution{U.S. Army Research Laboratory}
  \city{Adelphi, MD}
  \country{United States}}
\email{wesley.a.suttle.ctr@army.mil}

\author{Brian M. Sadler}
\affiliation{
  \institution{U.S. Army Research Laboratory}
  \city{Adelphi, MD}
  \country{United States}}
\email{brian.m.sadler6.civ@army.mil}

\begin{abstract}
    Deceptive path planning (DPP) is the problem of designing a path that hides its true goal from an outside observer.
    Existing methods for DPP rely on unrealistic assumptions, such as global state observability and perfect model knowledge, and are typically problem-specific, meaning that even minor changes to a previously solved problem can force expensive computation of an entirely new solution. Given these drawbacks, such methods do not generalize to unseen problem instances, lack scalability to realistic problem sizes, and preclude both on-the-fly tunability of deception levels and real-time adaptivity to changing environments.
    In this paper, we propose a reinforcement learning (RL)-based scheme for training policies to perform DPP over arbitrary weighted graphs that overcomes these issues. The core of our approach is the introduction of a local perception model for the agent, a new state space representation distilling the key components of the DPP problem, the use of graph neural network-based policies to facilitate generalization and scaling, and the introduction of new deception bonuses that translate the deception objectives of classical methods to the RL setting.
    Through extensive experimentation we show that, without additional fine-tuning, at test time the resulting policies successfully generalize, scale, enjoy tunable levels of deception, and adapt in real-time to changes in the environment.
\end{abstract}


\newcommand{\BibTeX}{\rm B\kern-.05em{\sc i\kern-.025em b}\kern-.08em\TeX}

\begin{document}


\pagestyle{fancy}
\fancyhead{}


\maketitle


\section{Introduction}

The capacity for deception is an indicator of intelligence \cite{turing1950mind}. Understanding how to deploy and counteract deceptiveness is critical to success in human affairs as diverse as business \cite{chelliah2018deception}, sports \cite{jackson2019deception}, and war \cite{tzu2008art}. Deception between humans has been widely studied in psychology \cite{hyman1989psychology}, economics \cite{gneezy2005deception}, and philosophy \cite{fallis2010lying}, yet the study of deception at the intersection of human and machine intelligence has recently experienced a surge of interest in the planning and artificial intelligence (AI) communities due to the rapid development of AI and autonomy and the increasing complexity of human-AI interaction.

\begin{figure}
    \vspace{1cm}
    \includegraphics[width=\linewidth]{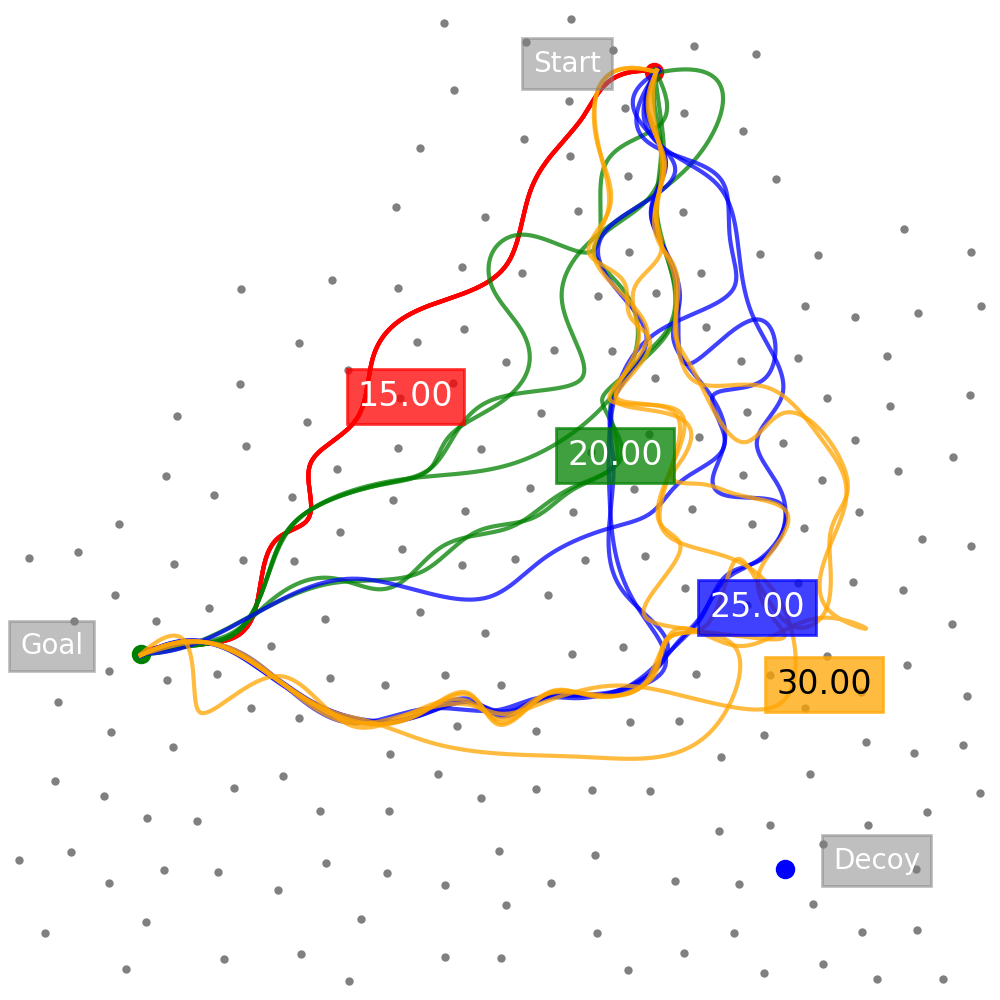}
    \caption{After training on only six small gridworld problems, our GNN-equipped RL agent is able to perform tunably deceptive navigation through a never-before-seen, continuous forest environment using only local perception. Deceptiveness is achieved through exaggeration towards a decoy goal and is tuned by allowing the agent 15, 20, 25, and 30 additional steps of ``time-to-deceive'' before reaching the goal.}
    \label{fig:exag_n10_urgency}
\end{figure}

One particularly active problem at the intersection of deception and autonomy is that of deceptive path planning (DPP): designing a path from a starting location to a goal location that hides the agent's true goal from an external, potentially adversarial observer. The community has produced a range of methods for solving this problem, including classical planning- and control-based methods \cite{masters2017deceptive, ornik2018deception, savas2022deceptive} as well as efforts at reinforcement learning-based approaches \cite{liu2021deceptive, Lewis_Miller_2023}.
Unfortunately, these methods all suffer from some combination of the following: the need for perfect knowledge of the environment, lack of scalability to realistic problem sizes, excessive computational overhead, lack of generalizability to unseen problems, and/or lack of deceptiveness tunability. The reasons for these drawbacks are primarily due to: (i) the model knowledge requirements of classical planning-based methods and model-based RL methods, and (ii) the inflexible and problem-specific perception models assumed by previous works.

The field of RL \cite{sutton2018reinforcement} -- a suite of techniques for approximately solving sequential decision-making problems formulated as Markov decision processes (MDPs) \cite{Puterman_2014} -- has seen incredible growth in recent years. Methods such as deep Q-learning (DQN) \cite{mnih2015human}, deep deterministic policy gradient (DDPG) \cite{lillicrap2015continuous}, proximal policy optimization (PPO) \cite{schulman2017proximal}, and soft actor-critic (SAC) \cite{haarnoja2018soft} have achieved impressive performance on a wide range of challenging control problems. Importantly, these techniques are all model-free in that they require no prior knowledge of the underlying environment, instead learning purely through experience. 
Graph neural networks (GNNs) are a class of neural network architectures consisting of repeated composition of graph convolutions and pointwise nonlinearities. Due to their invariance, stability, and transferability properties \cite{ruiz2019invariance, gama2020stability, ruiz2023transferability}, GNNs are particularly well-suited to problems where generalization and scalability to unseen, large graphs are critical, and have notched impressive practical successes \cite{kipf2016semi, defferrard2016convolutional, bronstein2017geometric}.

In this paper, we propose a novel RL scheme that, after training on only a limited set of small and relatively simple problems, produces autonomous agents capable of performing tunably deceptive DPP in larger, more complex, previously unseen environments. This simultaneously addresses an open problem in the DPP literature and lays the groundwork for performing DPP in real-world scenarios. The key aspects of our approach that allow it to overcome drawbacks of previous methods are the use of model-free RL to dispense with issue (i) described above, and combination of a novel, graph-based agent perception model with the use of GNN policies and value functions to address (ii).

First, we provide a novel formulation of the DPP problem as performing deceptive path planning over arbitrary weighted graphs that is amenable to solution using RL-based methods and easily represented as a Gym-conformant environment \cite{brockman2016openai}. Unlike in previous DPP works, our formulation incorporates a perception model for the agent that captures the essential aspects of the DPP problem while requiring the agent has only local information. Importantly, our formulation uses novel deceptive reward bonuses capturing the primary deception metrics of exaggeration and ambiguity of previous works \cite{masters2017deceptive, ornik2018deception, savas2022deceptive}.

Second, we develop an RL-based scheme for training general-purpose policies for performing DPP. After training on a small training set of only six different graphs representing fairly simple gridworld environments, our method produces policies that are able to generalize to significantly larger, more complicated environments and that can be tuned on-the-fly to display varying levels of deceptiveness. The key to our method is the use of GNN-based policies and value functions over the local neighborhood subgraph presented by the agent's local perception model, enabling the generalization ability and scalability that we see in our experiments. 

Finally, we experimentally validate the effectiveness of our approach on a range of discrete gridworld navigation problems as well as a continuous motion forest navigation problem. First, we present learning curves illustrating the effectiveness of our training scheme on a limited training set of six discrete DPP environments. Next, we present results demonstrating the scalability and generalization abilities of the resulting policies. We also perform ablation studies examining the effect of varying GNN model architectures, various hyperparameters, and other specifics of the training scheme. Finally, we provide experiments demonstrating the applicability of the trained policies to a continuous motion navigation problem as well as their tunable deceptiveness. Our experimental framework is publicly available at \cite{dpp_github}.
\section{Related Work}

Autonomous deception has been studied from a variety of perspectives, including detection of \cite{santos2009deception} and robustness to \cite{huang2019deceptive} deception, deceptive robotic motion planning \cite{dragan2014analysis, nichols2022adversarial}, game theory \cite{wagner2011acting, huang2021dynamic, rostobaya2023deception}, and even theory of mind \cite{sarkadi2019modelling}. DPP in particular was studied in \cite{masters2017deceptive}, which develops the simulation (also known as exaggeration) and dissimulation (also known as ambiguity) deception metrics, provides a formal characterization of the DPP problem, and proposes solution heuristics. This work was later generalized to multi-stage planning problems in \cite{price2023domain}. Though useful, \cite{masters2017deceptive, price2023domain} rely on perfect knowledge of the system model, the solutions proposed are heuristic, and the solutions do not generalize to previously unseen problems.

In \cite{ornik2018deception}, a formulation for general deception problems building on the MDP formalism is developed, yielding linear programming-based solution methods for obtaining optimal policies for performing deceptive planning. Extensions of \cite{ornik2018deception} include \cite{karabag2019optimal} to the problem of supervisory control, \cite{savas2022planning} to deceptive resource allocation, and \cite{savas2022deceptive} to the DPP problem. Most relevant for our work is naturally \cite{savas2022deceptive}, which combines an observer prediction model based on the principle of maximum entropy \cite{ziebart2008maximum, ziebart2009planning, ziebart2010modeling} with the approach of \cite{ornik2018deception} to obtain optimal policies for DPP. Like \cite{masters2017deceptive}, perfect knowledge of the system model is required and the resulting policies do not transfer to previously unseen problems. Due to the fact that its DPP formulation is MDP-based, however, \cite{savas2022deceptive} provides an important starting point for RL methods that can overcome these issues.

RL methods for DPP have seen limited attention in the literature. The work \cite{liu2021deceptive} builds on \cite{ornik2018deception} to develop an RL-based approach to privacy-preserving, deceptive planning, of which DPP is a special case. The resulting methods are privacy-preserving in that they make it difficult for an outside observer to infer the reward function of the learning agent and deceptive in that they leverage the notion of ambiguity developed in \cite{masters2017deceptive, ornik2018deception}. Like \cite{masters2017deceptive, ornik2018deception}, however, the policies learned are problem-specific and do not generalize to previous unseen problems.
Unlike \cite{liu2021deceptive}, which uses model-based value iteration as a core element of its approach, \cite{Lewis_Miller_2023} proposes a model-free RL method for the problem considered in \cite{liu2021deceptive}. Their method performs comparably with that from \cite{liu2021deceptive}, but differs from that work in that the Q functions for the candidate reward functions are learned instead of assumed to be known. Like \cite{masters2017deceptive, ornik2018deception, liu2021deceptive}, however, the learned policies are problem-specific and do not generalize. We address these shortcomings in this work.
\section{Problem Formulation} \label{sec:problem_formulation}

In this section we first describe the MDP-based deceptive path planning model of \cite{savas2022deceptive}. This introduces the fundamentals of the DPP problem setting, including the observer model and mathematical representation of deception that will be critical in what follows. We subsequently present our novel, RL-focused alternative.

\subsection{Classical Model} \label{subsec:classical_model}

There are three components to the DPP model studied in \cite{savas2022deceptive}: the underlying MDP used to capture the underlying navigation problem, the observer model based on this underlying MDP, and the mathematical representation of deception with respect to this observer model. We now describe these components.

\subsubsection{Navigation Model} Let an MDP $\mc{M} = (\mc{S}, \mc{A}, P, s_1, c, \gamma_c)$ be given, where $\mc{S}$ is the state space, $\mc{A}$ the action space, $P : \mc{S} \times \mc{A} \rightarrow \Delta(\mc{A})$ is the transition probability kernel mapping state-action pairs to probability distributions $\Delta(\mc{A})$ over $\mc{A}$, $s_1 \in \mc{S}$ is the initial state, $c : \mc{S} \times \mc{A} \rightarrow \mathbb{R}$ is the cost function, and $\gamma_c \in (0, 1)$ is a discount factor. A policy $\pi : \mc{S} \rightarrow \Delta(\mc{A})$ maps states to probability distributions over $\mc{A}$. At a given timestep $t \in \mathbb{N}$, an agent equipped with policy $\pi$ will take an action $a_t \sim \pi(\cdot | s_t)$, receive a cost $c_t = c(s_t, a_t)$, and transition to state $s_{t+1} \sim P(\cdot | s_t, a_t)$. Let $\mc{G} \subset \mc{S}$ denote the set of potential goals and let $G^* \in \mc{G}$ denote the agent's true goal. Furthermore, let $\text{Pr}^{\pi}(Reach(G))$ denote the probability that following policy $\pi$ will eventually lead the agent into state $G \in \mc{G}$, and let $R_{max}(G) = \max_{\pi} \text{Pr}^{\pi}(Reach(G))$ denote the maximum probability of reaching $G$ under any policy. In the standard MDP setting, the goal of the agent would be to find a policy $\pi^*$ such that $\text{Pr}^{\pi^*}(Reach(G)) = R_{max}(G)$ while incurring minimal cost. Finally, in what follows we will use $\zeta = (s_1, a_1, s_2, a_2, \ldots)$ to denote a trajectory of state-action pairs, and $\zeta_{1:T}$ a partial trajectory of length $T \in \mathbb{N}$.

\subsubsection{Observer Model} Building on the MDP navigation model described above, \cite{savas2022deceptive} proposes an observer model enabling the agent to predict the observer's belief about its true goal, given its partial trajectory $\zeta_{1:T}$ up to time $T$. Specifically, the model provides a probability distribution $P(G | \zeta_{1:T})$ over all possible goals $g \in \mc{G}$. This is accomplished by assuming that, given that the agent's goal is $G \in \mc{G}$, the observer expects the agent to rationally seek the lowest-cost path with respect to cost function $c$ while accommodating for a level of inefficiency specified by a scalar parameter $\alpha > 0$, where the agent becomes completely rational and efficient as $\alpha \downarrow 0$ and inefficiency is an increasing function of $\alpha$.
As shown in \cite{savas2022deceptive}, the observer model under this assumption is given by
\begin{equation}
    \text{Pr}(G | \zeta_{1:T}) \approx \frac{ e^{V_G(s_T) - V_G(s_1)} \text{Pr}(G) }{ \sum_{G' \in \mc{G}} e^{V_{G'}(s_T) - V_{G'}(s_1)} \text{Pr}(G') }, \label{eqn:observer_model}
\end{equation}
where $V_G$ is obtained using the equations
\begin{align}
    Q_G(s, a) &= - c(s, a) + \gamma_c \sum_{s' \in \mc{S}} P(s' | s, a) V_G(s') \label{eqn:softmax_Q} \\
    V_G(s) &= \alpha \log \sum_{a \in \mc{A}} e^{Q_G(s, a) / \alpha}. \label{eqn:softmax_V}
\end{align}
Given knowledge of $P$ and $c$, $V_G$ can be computed via softmax value iteration \cite{ziebart2008maximum}. As discussed in \cite{savas2022deceptive}, since \eqref{eqn:observer_model} is only a function of $s_1$ and $s_T$, the entire observer prediction model \eqref{eqn:observer_model} can be obtained offline by computing $V_G(s)$, for all $G \in \mc{G}, s \in \mc{S}$, which can be achieved by running softmax value iteration a total of $|\mc{G}|$ times.

\subsubsection{Deception Representation} The two most common types of deception considered in the DPP literature are exaggeration and ambiguity. In the exaggeration setting, the agent misleads the observer about its true goal by navigating towards a decoy goal instead. This is naturally represented using the observation model detailed above by the function
\begin{equation}
    f(s_t, a_t) = 1 + \text{Pr}(G^* | \zeta_{1:t} ) - \max_{G \in \mc{G} \setminus G^*} \text{Pr}(G | \zeta_{1:t} ). \label{eqn:classical_exaggeration}
\end{equation}
Since $\text{Pr}(G | \zeta_{1:T})$ tends to become larger as $\zeta_{1:T}$ approaches $G$, finding a policy that minimizes \eqref{eqn:classical_exaggeration} subject to the constraint that $G^*$ is eventually reached clearly encourages paths that exaggerate the likelihood of entering a decoy goal state by passing close by it. To see this, notice that the objective achieves its minimum when $\pi$ is such that $\text{Pr}(G^* | \zeta_{1:t} ) = 0$ and $\max_{G \in \mc{G} \setminus G^*} \text{Pr}(G | \zeta_{1:t} ) = 1$, and is otherwise increasing in $\text{Pr}(G^* | \zeta_{1:t} ) - \max_{G \in \mc{G} \setminus G^*} \text{Pr}(G | \zeta_{1:t} )$.

In the ambiguity setting, the agent misleads the observer by selecting paths that remain noncommittal regarding the true goal for as long as possible. This is naturally represented by the function taking the value
\begin{equation}
    f(s_t, a_t) = \sum_{G \in \mc{G}} \sum_{G' \in \mc{G}} | P(G | \zeta_{1:t}) - P(G' | \zeta_{1:t}) | \label{eqn:classical_ambiguity}
\end{equation}
when $s_t \in \mc{S} \setminus \mc{G}$ and $0$ when $s_t \in \mc{G}$. A policy minimizing \eqref{eqn:classical_ambiguity} will tend to generate trajectories for which the observer assigns all goals equal probability, rendering the true goal $G^*$ indistinguishable from the other elements in $\mc{G}$.

\subsection{Our Model} \label{subsec:our_model}

We now describe our DPP model, which is the setting of the GNN-equipped RL training scheme detailed in the following section. As in \S\ref{subsec:classical_model}, our objective remains to generate movement towards the true goal $G^* \in \mc{G}$ while deceiving the observer as to the identity of $G^*$. However, unlike performing deception under the model described in \S\ref{subsec:classical_model}, which is solved in \cite{savas2022deceptive} in an offline manner using linear programming, we aim for a problem formulation that can be solved in an online manner using RL. As we will see in \S\S\ref{sec:method}-\ref{sec:experiments}, this paves the way for learning policies that generalize to unseen problem instances and are scalable and tunable, unlike the solutions generated by existing methods. To achieve this, we first provide a novel, graph-based state model, then define suitable deception bonuses for the RL setting. These deception bonuses will subsequently be used in the construction of rewards for training policies to perform DPP using RL in \S\ref{sec:method}.

\subsubsection{Graph-based State Model} \label{subsubsec:graph_model}
At a fundamental level, we represent the environment to be navigated as an undirected, weighted graph $\mc{H} = (\mc{S}, \mc{A}, c)$, where $\mc{S}$ is the set of states, or nodes, in the graph, $\mc{A} \subset \mc{S} \times \mc{S}$ is the set of edges representing accessibility between nodes, and $c : \mc{A} \rightarrow \mathbb{R}$ is the edge weight mapping. For a fixed, prespecified integer $k \geq 0$, define the $k$-hop neighborhood of $s \in \mc{S}$ by $\mc{N}_k(s) = \{ s' \in \mc{S} \ | \ d_{\mc{H}}(s, s') \leq k \}$, where $d_{\mc{H}}(s, s')$ is the shortest number of edges that must be traversed to move from $s$ to $s'$ in $\mc{H}$. These local neighborhoods form the basis for our agent's perception model: when the agent is at state $s$, the region $\mc{N}_k(s)$, or visibility graph, is visible to it. We furthermore associate with each $s_t$ the vector of node attributes
\begin{equation}
    \left[ \bm{1}_{visited}(s_t) \hspace{2mm} d_c(s_t, G_1) \hspace{2mm} \ldots \hspace{2mm} d_c(s_t, G_{|\mc{G}|}) \hspace{2mm} T_{max} - t \right]^T, \label{eqn:node_attributes}
\end{equation}
where $\bm{1}_{visited}(s_t) = 1$ if the agent has previously visited $s_t$ and $0$ otherwise, $d_c(s_t, G_k)$ is the minimum distance path from $s_t$ to goal $G_k \in \mc{G}$, for $k \in \{1, \ldots, |\mc{G}| \}$, and $T_{max}$ is a user-specified maximum number of allowable steps by which the agent should reach $G^*$ during an episode. We will henceforth abuse notation and identify $s_t$ with \eqref{eqn:node_attributes} and the visibility graph $\mc{N}_k(s_t)$ with the set of node attribute vectors corresponding to all $s \in \mc{N}_k(s_t)$.

\subsubsection{Deception Representation} \label{subsubsec:our_model_deception}
We now formulate the deception bonuses leveraged in \S\ref{sec:method} to train policies for DPP. As in previous works, we consider the exaggeration and ambiguity notions of deception. We consider the same observer model-reliant notion of exaggeration developed in \cite{savas2022deceptive}, which we convert to a reward by modifying \eqref{eqn:classical_exaggeration} to obtain
\begin{equation}
    r_e(\zeta_{1:t}) = \max_{G \in \mc{G} \setminus G^*} \text{Pr}(G | \zeta_{1:t}) - \text{Pr}(G^* | \zeta_{1:t}). \label{eqn:our_exaggeration}
\end{equation}

For ambiguity, we found that the straightforward modification $r_a(\zeta_{1:t}) = -f_a(\zeta_{1:t})$ of \eqref{eqn:classical_ambiguity} led to training difficulties in an RL setting. We therefore instead define ambiguity using the function
\begin{equation}
    r_a(\zeta_{1:t}) = \sum_{G \in \mc{G}} {\left(1 - \frac{|d_c(s_t, G) - d_c(s_t, G^*)|}{d_c(G, G^*)}\right)} \label{eqn:our_ambiguity}
\end{equation}
Intuitively, this represents the disparity between proximity to decoy goals $G \in \mc{G}$ and the true goal $G^*$, normalized by the distance $d_c(G, G^*)$, which provides an upper bound on the discrepancy $|d_c(s_t, G) - d_c(s_t, G^*)|$. This enables the RL approach detailed in the following section to successfully learn ambiguous path planning.
%
%

\begin{figure}
     \centering
     \begin{subfigure}[b]{0.23\textwidth}
        \centering
        \includegraphics[width=\linewidth]{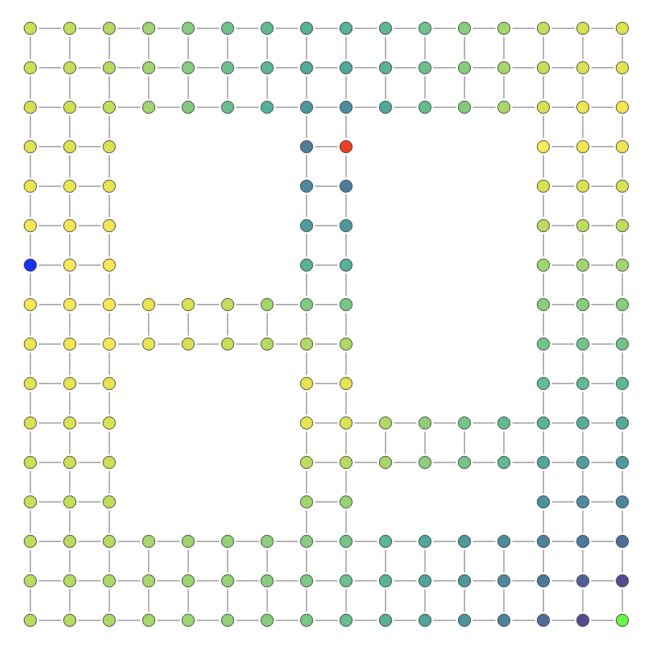}
        \caption{Classical ambiguity}
        \label{fig:ambiguity_heatmap_old}
     \end{subfigure}
     \hfill
     \begin{subfigure}[b]{0.23\textwidth}
        \centering
        \includegraphics[width=\linewidth]{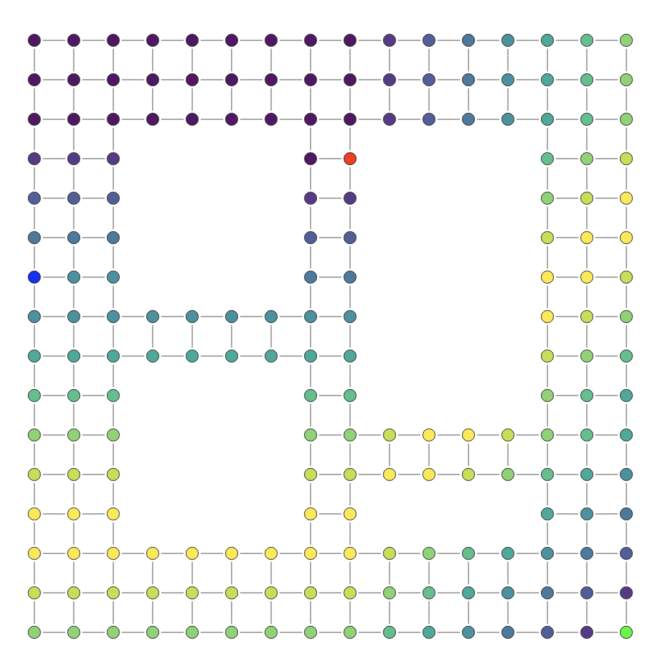}
        \caption{Proposed ambiguity}
        \label{fig:ambiguity_heatmap_new}
     \end{subfigure}
     \caption{Comparison of classical ambiguity from \cite{savas2022deceptive} with our \eqref{eqn:our_ambiguity}. Colors denote: \textbf{\textcolor{blue}{start position}}, \textbf{\textcolor{ForestGreen}{true goal}}, \textbf{\textcolor{red}{decoy goal}}.}
     \label{fig:ambiguity_justification}
\end{figure}

As illustrated in Figure \ref{fig:ambiguity_justification}, the original ambiguity metric \eqref{eqn:classical_ambiguity} creates a reward penalty around each goal, while the replacement \eqref{eqn:our_ambiguity} that we propose creates a reward bonus in the “tightrope” region that can be seen between the two goals. This is likely due to the softmax value calculation of equations \eqref{eqn:softmax_Q}, \eqref{eqn:softmax_V}. As the distance from either goal grows, the absolute difference in the value function decreases, which weakens the reward signal at further distances from the goal. This may have worked for the linear programming method of \cite{savas2022deceptive} because it had access to the whole path as an optimizer; however, when sampling trajectories in the RL setting, our alternative formulation guides the agent towards an ambiguous path more effectively.
%
    

\section{Method} \label{sec:method}

In this section we describe our RL-based method for solving the DPP problem based on the model described in \S\ref{subsec:our_model}. The overall goal is to learn a policy $\pi$ that, given an arbitrary graph that we wish to perform DPP over, is able at each timestep $t$ to select an edge $a_t = (s_t, s) \in \mc{A}$ to traverse, based on its local observation $\mc{N}_k(s_t)$, such that deception is maximized while $G^*$ is reached within a user-specified number of steps $T_{max}$.
To achieve this, we start by designing the deceptive reward function that we use in training, which is based on the deception bonuses defined in \S\ref{subsubsec:our_model_deception}. Next, we develop the GNN architecture essential to the generalization and scalability properties of our method. Finally, we discuss the training scheme used to train the flexible RL agents evaluated in the experiments of \S\ref{sec:experiments}.

\subsection{Deceptive Rewards} \label{subsec:reward_function}

To address the DPP problem using RL, we ultimately need to define a reward function that balances operating deceptively with respect to the deception bonuses of \S\ref{subsubsec:our_model_deception} with reaching $G^*$ in a timely manner. For a given deceptive bonus $r \in \{ r_e, r_a \}$ from \S\ref{subsubsec:our_model_deception}, we achieve this using the following reward function:
\begin{equation*}
    R_t = R(\zeta_{1:t}) = \left\{
    \begin{array}{lr}
        r(\zeta_{1:t}) &\text{if } s_t \notin \zeta_{1:t-1} \\
        1 &\text{if } s_t = G^* \\
        -1 &\text{if } t > T_{max} \\
        0 &\text{otherwise}
    \end{array}
\right\}
\end{equation*}
This function penalizes going past the time limit $T_{max}$ with $-1$, rewards reaching the goal with $+1$, and discourages paths containing cycles by only dispensing the ``deceptive bonus" $r(\zeta_{1:t})$ the first time an agent reaches $s_t$. In order to teach the agent to prioritize reaching the goal over acting deceptively, we also retroactively nullify deceptive rewards if the goal was not reached before $T_{max}$. 
%
%
With this reward in hand, our agents will be trained to find a policy $\pi$ maximizing the objective 
%
%
\begin{equation}
    J(\pi) = \mathbb{E}_{\pi} \left[ \sum_{t=1}^T \gamma^{t-1} r(\zeta_{1:t}) \right], \label{eqn:discounted_reward}
\end{equation}
where $\gamma \in (0, 1)$ is a user-specified discount factor and the horizon is given by $T = \min \{ \tau \ | \ G^* \in \zeta_{1:\tau} \}$,
%
%
the first timestep at which the agent reaches $G^*$.

\subsection{Graph Neural Network Architecture} \label{subsec:gnn_architecture}

\begin{figure}
    \includegraphics[width=\linewidth]{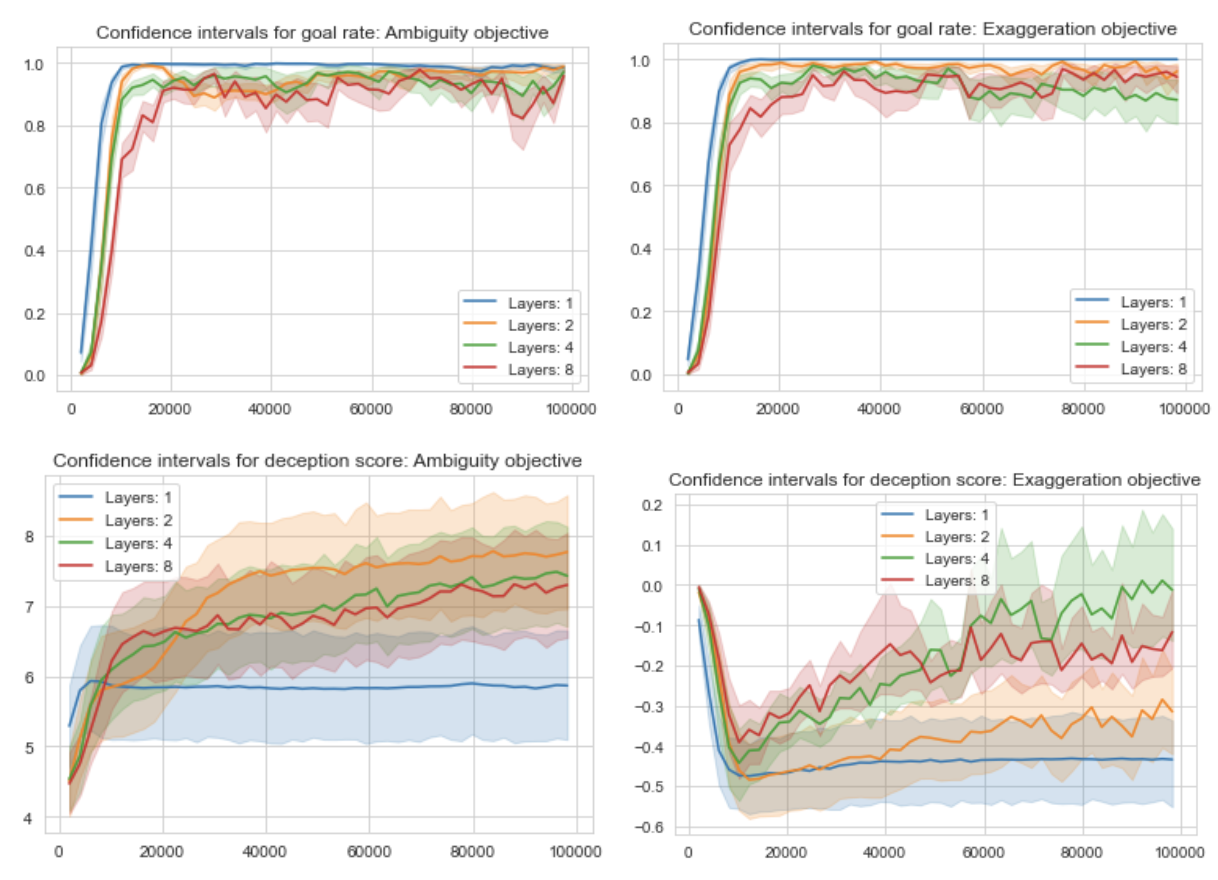}
    \caption{Learning curves for various $k$, the number of GNN layers and the radius of the agent's $k$-hop neighborhood. Curves present mean and 95\% confidence intervals over five independent replications. We found that for ambiguity, one layer is not enough to effectively act deceptively, while performance peaks on validation data at two layers before dropping off again for four and eight layers. For exaggeration-tuned behavior, the optimal $k$ is four, where one layer is again not enough to act deceptively, two has a slight improvement in performance, and four has the best performance.}
    \label{fig:nbhd_radius}
\end{figure}

\begin{figure}
     \centering
     \begin{subfigure}[b]{0.22\textwidth}
         \centering
         \includegraphics[width=\textwidth]{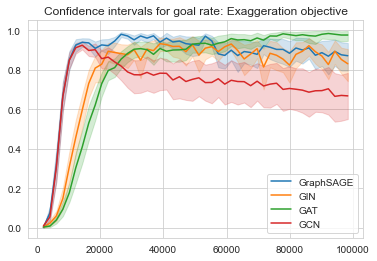}
         \caption{goal rate for exaggeration}
         \label{fig:arch_comparison_ex_goalrate}
     \end{subfigure}
     \hfill
     \begin{subfigure}[b]{0.22\textwidth}
         \centering
         \includegraphics[width=\textwidth]{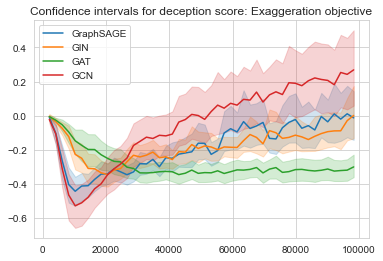}
         \caption{exaggeration deceptiveness}
         \label{fig:arch_comparison_ex_deceptiveness}
     \end{subfigure}

     \vspace{2mm}
     \begin{subfigure}[b]{0.22\textwidth}
         \centering
         \includegraphics[width=\textwidth]{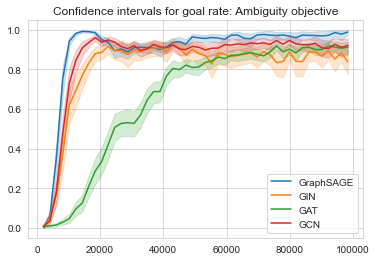}
         \caption{goal rate for ambiguity}
         \label{fig:arch_comparison_am_goalrate}
     \end{subfigure}
     \begin{subfigure}[b]{0.22\textwidth}
         \centering
         \includegraphics[width=\textwidth]{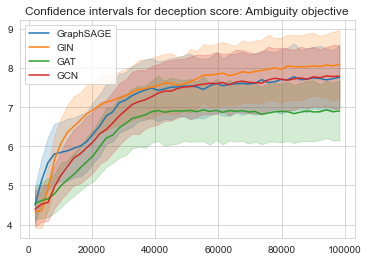}
         \caption{ambiguity deceptiveness}
         \label{fig:ex100}
     \end{subfigure}
        \caption{GNN architecture comparison. Curves present mean and 95\% confidence intervals over five independent replications. For exaggeration-tuned behavior, we found a trade-off between deceptiveness and path efficiency; this is to be expected, as extra bias in the path towards a decoy goal adds extra distance compared to the baseline shortest path. For ambiguity, we found that GraphSAGE and graph isomorphism networks were able to reach the goal under the time limit reasonably frequently, while also maximizing the level of ambiguity in the generated path. The graph attention network performed similarly for ambiguity, but exhibited an inability balance acting deceptively with reaching the goal.}
        \label{fig:model_arch_comparison}
\end{figure}

We use graph neural networks (GNNs) to infer the best deceptive action given the subgraph $\mc{N}_k(s_t)$ of the environment visible to the agent at each timestep $t$, as GNNs have been shown to exhibit high generalizability and applicability to complex environments \cite{chenning2021collision, dai2019graphexploration}.
A GNN operates by gradually introducing context from a node's surroundings into an internal representation by the model. This is commonly achieved by defining a message-passing operation between neighboring nodes \cite{kipf2017graphconvolutionalnetworks, velickovic2018graphattentionnetworks, xu2019graphisomorphismnetworks, hamilton2018graphsage}.
The message-passing mechanism that we use is GraphSAGE \cite{hamilton2018graphsage}, which incorporates a specialized max-pooling operator and explicitly separated node and neighbor embeddings. 
%
%
We focus on GraphSAGE in this work, but results indicate that graph isomorphism networks also give good results (see the ablation study in Figure \ref{fig:model_arch_comparison}). We use a $k$-layer GraphSAGE network, which is preceded by a linear layer to project the 4-dimensional node attribute vector associated with each $s \in \mc{N}_k(s_t)$ at time $t$ into a 64-dimensional intermediate feature vector space. The feature vector associated with state $s$ is then updated by sampling up to $k$ neighbors from $\mc{N}_k(s)$, and the process repeats for each layer of the network.
We experiment with model architectures having $k \in \{1, 2, 3, 4\}$ layers, using $64$ dimensions for feature vector sizes throughout all layers (see Figure \ref{fig:nbhd_radius}).

\subsection{Training Scheme} \label{subsec:training_procedure}

\begin{figure}
    \includegraphics[width=\linewidth]{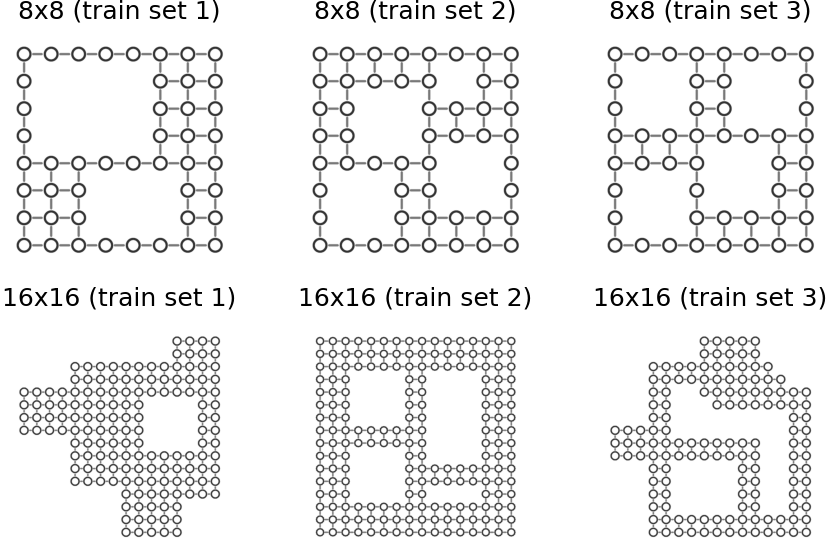}
    \caption{The set of grid worlds used in training. We considered three $8\times8$ and three $16\times16$ topologies and found that this was sufficient for generalization.}
    \label{fig:training_gridworlds}
\end{figure}

Using the reward and GNN architectures detailed above we trained two policies, one each for exaggeration and ambiguity, on a small but representative set of training environments: three $8 \times 8$ gridworlds and three $16 \times 16$ gridworlds (see Figure \ref{fig:training_gridworlds}). For simplicity, in this work we focused on the single-decoy goal setting, we set the efficiency parameter $\alpha = 1$ in \eqref{eqn:softmax_V} in our exaggeration reward $r_e$, and we assumed that the edge weights $c$ of the graph $\mc{H}$ defined in \S\ref{subsubsec:graph_model} were all 1. We trained using PPO with discount factor $\gamma = 0.99$ for a total of $98304$ episodes using the Adam optimizer \cite{kingma2017adam} with learning rate and weight decay parameters both set to $0.0004$. At the beginning of each episode we randomly sampled the environment, start position $s_1$, true goal $G^*$, decoy goal $G$, and time limit $T_{max}$. Environments we sampled uniformly at random, true and decoy goals were selected uniformly at random without replacement, and $T_{max}$ was sampled according to $T_{max} \sim \text{Uniform}(d_c(s_1, G^*), d_c(s_1, G) + d_c(G, G^*))$. By training across a wide variety of $T_{max}$ values, we found that we could control the level of ``urgency'' the model used when balancing reaching $G^*$ with behaving deceptively.
%
%
%
%
%

\section{Gridworld Experiments} \label{sec:experiments}

In this section we present experiments illustrating the performance of agents trained using the methods described in \S\ref{sec:method} on previously unseen gridworld environments. The experiments demonstrate the generalization and scalability of the DPP policies as well as the tunability of their level of deceptiveness.

%

\subsection{Generalization and Scalability} \label{subsec:generalization_scalability}

\begin{figure}
     \centering
     \begin{subfigure}[t]{0.22\textwidth}
        \centering
        \includegraphics[width=\linewidth]{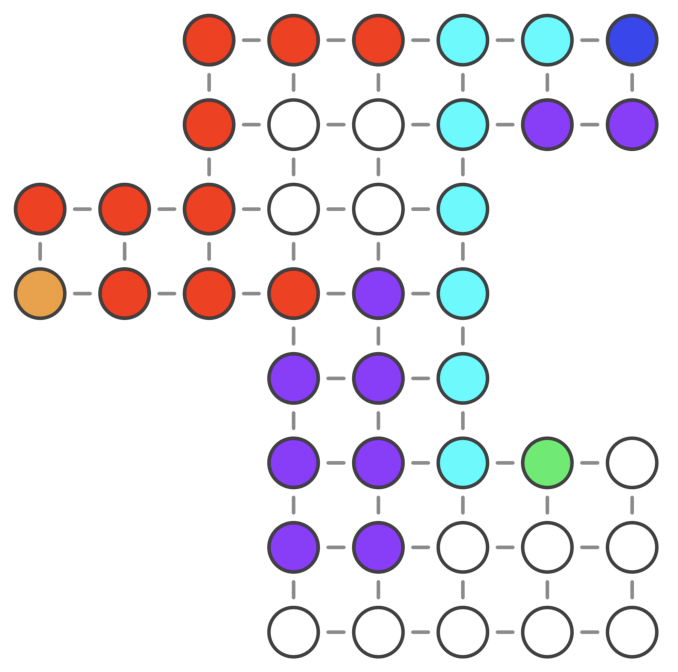}
        \caption{$8 \times 8$ graph}\label{fig:ex8}
     \end{subfigure}
     \hfill
     \begin{subfigure}[t]{0.22\textwidth}
        \centering
        \includegraphics[width=\linewidth]{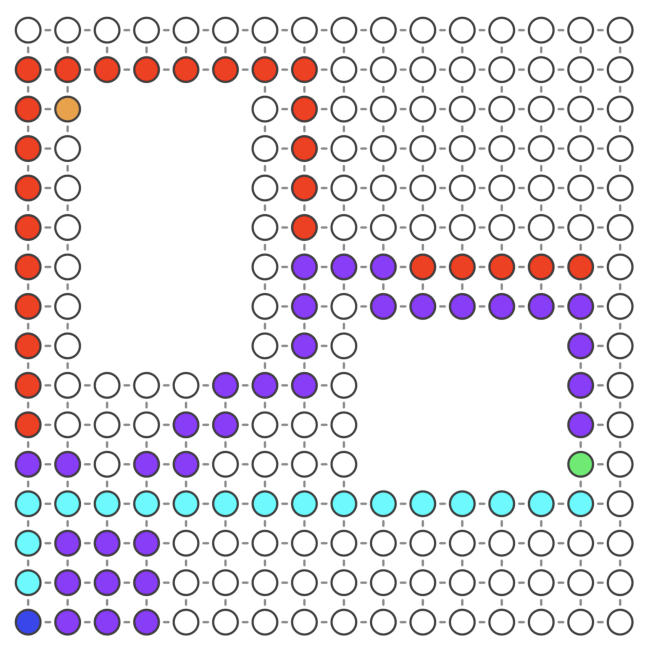}
        \caption{$16 \times 16$ graph}\label{fig:ex16}
     \end{subfigure}
     
     \vspace{2mm}
     \begin{subfigure}[t]{0.22\textwidth}
        \centering
        \includegraphics[width=\linewidth]{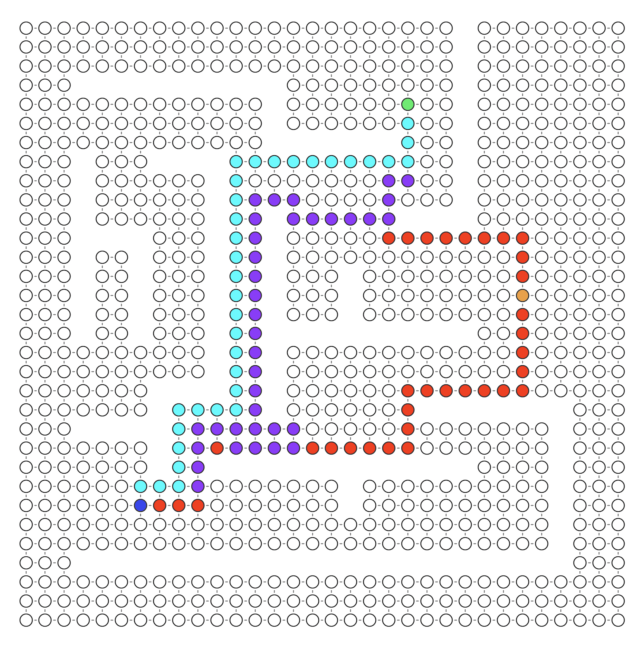}
        \caption{$32 \times 32$ graph}\label{fig:ex32}
     \end{subfigure}
     \hfill
     \begin{subfigure}[t]{0.22\textwidth}
        \centering
        \includegraphics[width=\linewidth]{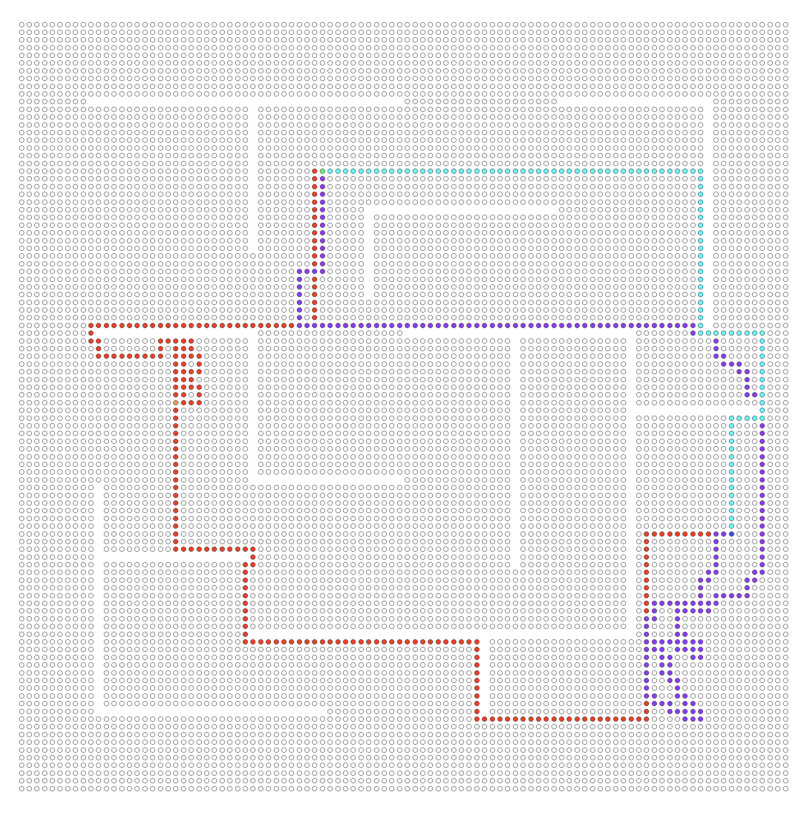}
        \caption{$100 \times 100$ graph}\label{fig:ex100}
     \end{subfigure}
        \caption{Generalization to unseen graphs and scalability to larger graphs. Colors denote: \textbf{\textcolor{ProcessBlue}{shortest path}}, \textcolor{violet}{\textbf{ambiguous path}}, \textcolor{red}{\textbf{exaggerated path}}, \textbf{\textcolor{blue}{start position}}, \textbf{\textcolor{ForestGreen}{true goal}}, \textbf{\textcolor{orange}{decoy goal}}.}
        \label{fig:scalability}
\end{figure}

The first set of experiments, presented in Figure \ref{fig:scalability}, illustrates the ability of our DPP policies to generalize to previously unseen problems (\ref{fig:ex8} and \ref{fig:ex16}) and to scale to larger problems than those encountered encountered during training (\ref{fig:ex32} and \ref{fig:ex100}). The trajectories pictured were generated by simply applying the exaggeration and ambiguity policies trained during the procedure depicted in Figure \ref{fig:nbhd_radius} with a perception radius of $k=4$ to the gridworlds in Figure \ref{fig:scalability}. We manually selected the start and goal locations in a way that we believed would best highlight the model's deceptive capabilities, then gave the agent $T_{max} = 1.5 \cdot d_c(s_1, G^*)$ steps to reach the goal.

The results pictured indicate that both policies are able to effectively balance the incentive to move towards the true goal with their respective deception incentives. In the case of the exaggeration policy, all trajectories pass by the decoy goal first before proceeding to the true goal. Similarly, the ambiguity policy generates trajectories that appear noncomittal for as long as possible. These results illustrate the basic effectiveness of our approach, but also highlight the power of the combination of our novel DPP model and RL-based training method: equipped with only local observability of $\mc{N}_k(s_t)$ at each timestep, the policies are able to design paths whose deceptiveness is only perceptible at a scale beyond that allowed by the agent's limited, local perception model. We hypothesize that this is due to the state model we propose (\S\ref{subsubsec:graph_model}), our use of GNNs (\S\ref{subsec:gnn_architecture}), and the fact that, because we varied start and goal positions in the training data (\S\ref{subsec:training_procedure}), the agents were forced to generalize to a variety of goal and decoy distances to plan over, resulting in more general, scalable policies.

\subsection{Deceptiveness Tunability} \label{subsec:tunability_experiments}

We next present experiments demonstrating that the deceptiveness level of a policy can be tuned without retraining. Tunability is achieved using the time constraint $T_{max}$. At any given time $T_{max}$ can be dynamically altered, and extending it beyond the minimum time needed for the shortest path to the goal from the current agent location introduces more deception into the policy.  Note that we have normalized time and steps, so that $T_{max}$ can be regarded as either a number of steps or time allotted. We find that by training the models over a variety of time limits as described in \S\ref{subsec:training_procedure}, the policies learn to reach the goal quickly when the time limit is too low for deception, and to take advantage of extra time to create deceptive behavior.

\begin{figure}[t]
     \centering
     \begin{subfigure}[t]{0.22\textwidth}
        \centering
        \includegraphics[width=\linewidth]{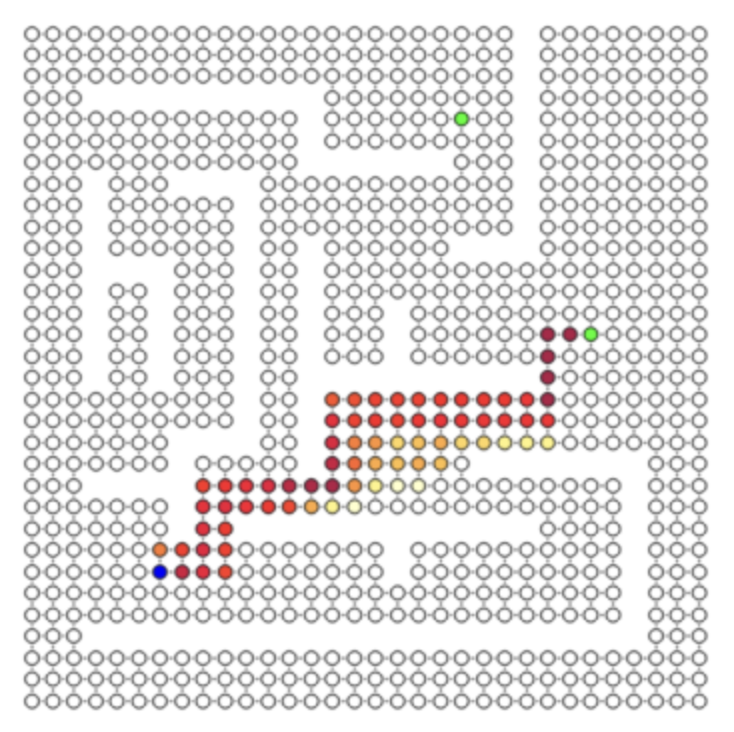}
        0 extra steps\label{fig:medium_ambiguity_0_extra}
     \end{subfigure}
     \hfill
     \begin{subfigure}[t]{0.22\textwidth}
        \centering
        \includegraphics[width=\linewidth]{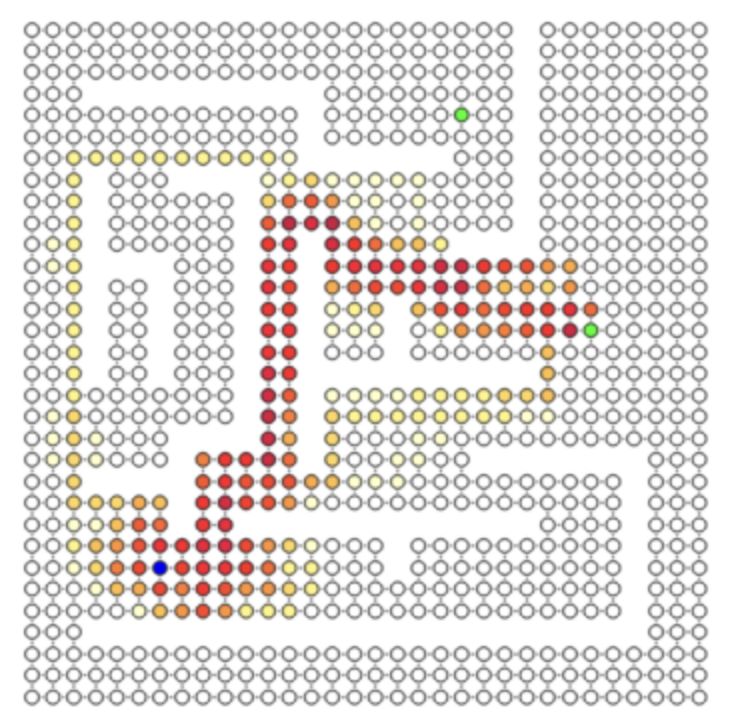}
        30 extra steps\label{fig:medium_ambiguity_30_extra}
     \end{subfigure}
     
     \vspace{2mm}
     \begin{subfigure}[t]{0.22\textwidth}
        \centering
        \includegraphics[width=\linewidth]{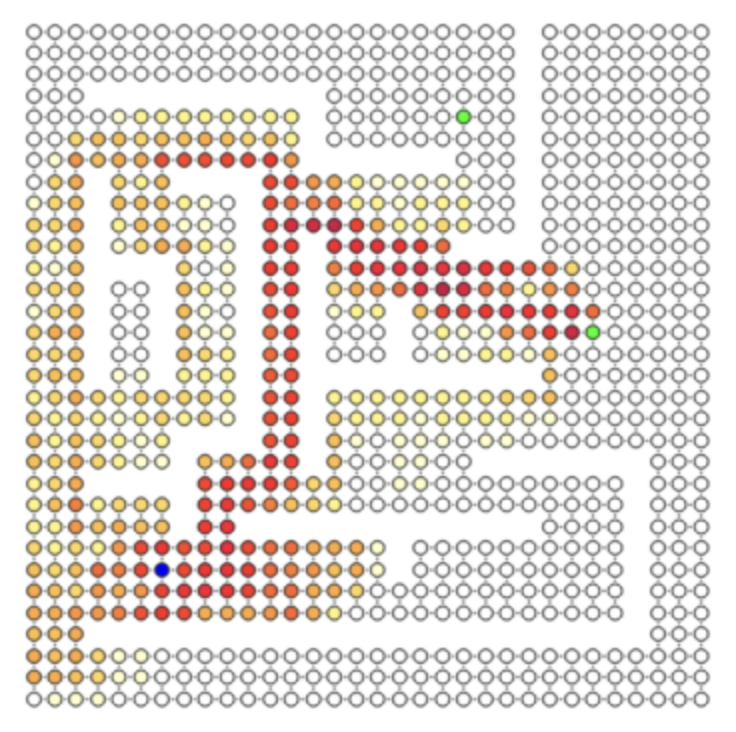}
        60 extra steps\label{fig:medium_amgibuity_60_extra}
     \end{subfigure}
     \hfill
     \begin{subfigure}[t]{0.22\textwidth}
        \centering
        \includegraphics[width=\linewidth]{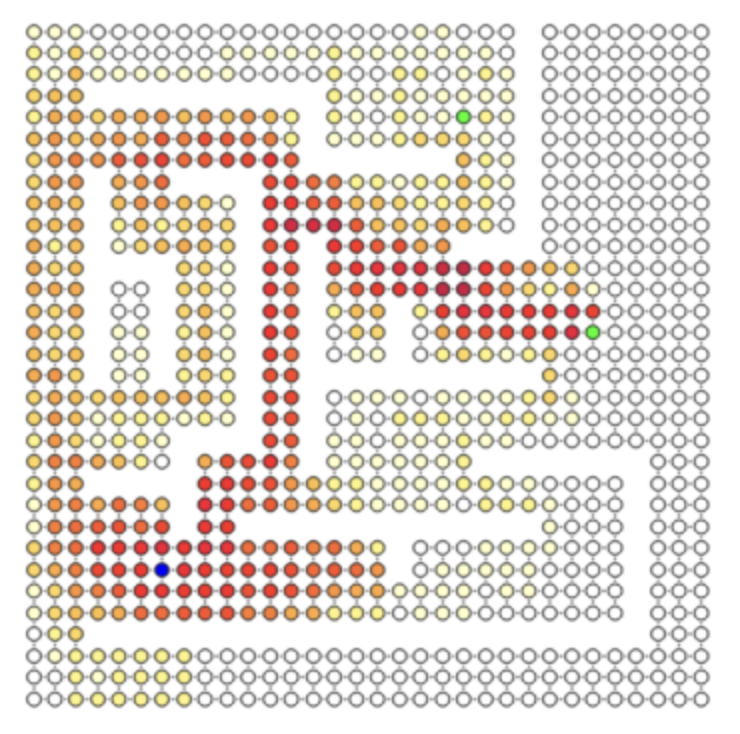}
        90 extra steps\label{fig:medium_ambiguity_90_extra}
     \end{subfigure}
        \caption{Deceptiveness tunability of exaggeration policy by adding extra steps to $T_{max}$. Heatmap generated using 32 trajectories. Darker colors indicate higher density. Colors denote: \textbf{\textcolor{blue}{start position}}, \textbf{\textcolor{ForestGreen}{true goal}}, \textbf{\textcolor{red}{decoy goal}}. At inference, the model opts for the shortest path to the goal when given $0$ extra steps to act deceptively, but when given $30$, $60$, or $90$ extra steps, the model adapts its behavior and the probability of increasingly ambiguous and distance paths increases.}
        \label{fig:tunability}
\end{figure}

The results of a tunability experiment for the exaggeration policy are shown in Figure \ref{fig:tunability}. Similar results hold for the ambiguity policy. Each of the heatmaps presented in Figure \ref{fig:tunability} summarizes 32 independent trajectories generated by the policy for a specific deceptiveness level. With 0 extra steps, $T_{max}$ is always equal to the minimum number of steps from the current agent position to the goal, and the policy seeks the minimum length path to the goal without regard to deception. Successive panels in Figure \ref{fig:tunability} correspond to allotting 30, 60, or 90 steps beyond the current minimum step length to the goal. As the extra step size grows the policy yields progressively more deceptive motion. We emphasize that this variable may be altered at any time by the agent, enabling online tunability of deception level without the need for replanning.

\section{Continuous Experiments} \label{sec:continuous_experiments}

All examples up to this point have been on discrete gridworlds and have used a discrete grid-step model. In this section we apply our framework to DPP on a continuous forest navigation problem. We first formulate the problem in a way that is amenable to the graph-based model developed in \S\ref{subsec:our_model}, then experimentally show that the same deceptive policies trained in \S\ref{sec:method} and evaluated in \S\ref{sec:experiments} can be directly applied to this problem without any fine-tuning.


\subsection{Continuous Forest Navigation Problem}

The continuous, 2-D ``forest'' environment that we propose is shown in Figure \ref{fig:exag_n10_urgency}.
%
%
The objective in this setting, as before, is to deploy the graph-inference deceptive motion policies from \S\ref{sec:method} to enable an agent to move to a true goal while employing a decoy goal. The key challenges are to formulate a graph representation that the policies can use to navigate through the forest as well as meaningful notions of the goal and decoy distances needed for deception.

\subsubsection{Graph Formulation} \label{subsubsec:continuous_graph_formulation}

Given a global map of the forest providing all tree positions, a reasonable way to construct the desired representation is to use Voronoi clustering to generate a set of graph nodes $\mc{S}$ and edges $\mc{A}$ to perform planning over. Voronoi clustering results in a set of polygonal regions, each containing one of the trees, such that they cover the 2-D region and every point on the plane belongs to the Voronoi region of the closest tree. Given such a Voronoi clustering, we can take vertices of the Voronoi regions to form the nodes $\mc{S}$ in the planning graph and the boundaries between neighboring Voronoi regions, or ``ridges'', to form edges $\mc{A}$ between the two vertices they correspond to. Denote the corresponding global planning graph by $\mc{H}_{glob} = (\mc{S}, \mc{A})$. We choose to use Voronoi regions to generate this graph structure because Voronoi ridges correspond to the set of locations at an equal distance between any two trees. Planning over $\mc{H}_{glob}$ thus provides a way to generate the safest possible paths through the forest, as any deviation from a Voronoi ridge would be unnecessarily close to one of the trees.

In realistic settings, the agent will not be equipped with a map of all tree positions, but will instead need to construct a local map of tree positions through, for example, vision or LIDAR. Assuming a perception disk of fixed radius centered at the agent location, we can perform a local approximation of the global Voronoi clustering scheme outlined above to obtain a local planning graph, $\mc{H}_{loc}$. In this setting, only trees contained within the perception disk around the agent's current location are considered when formulating the planning graph via Voronoi clustering, and the agent reasons over only the vertices and ridges contained within its perception disk. When the disk radius is large enough, however, for reasonably small values of the message-passing parameter $k$ in the GNN architecture considered in \S\ref{subsec:gnn_architecture} (such as $k=4$ used in our experiments below), the local neighborhood subgraphs $\mc{N}_k(s_t)$ that our policies use for planning will be similar in both $\mc{H}_{glob}$ and $\mc{H}_{loc}$.

\subsubsection{Deception Formulation} \label{subsubsec:continuous_deception_formulation}

To complete the problem specification for performing deception we assume that, for any given node of the planning graph $\mc{H}_{loc}$, the Euclidean distance from the node location to the goal and decoy are known. The Euclidean distance can be easily calculated based only on agent, goal, and decoy location, without regard to the obstacles (trees). Consequently, our approach does not require a map of tree positions a priori, relying instead on its local perception capabilities and local planning graph $\mc{H}_{loc}$. In the following results the same training and inference approach was used as in the prior examples and we did not fine-tune the policies to the new setting.

\subsection{Results}

In this final set of experiments, we evaluated the policies trained in \S\ref{sec:method} on the continuous forest navigation problem. We again highlight that, even though we originally trained the model in the gridworld setting, no fine-tuning was necessary for this experiment.



\subsubsection{Deceptiveness Tunability}

\begin{figure*}[htb]
     \centering
     \begin{subfigure}[t]{0.23\textwidth}
        \centering
		\includegraphics[width=\linewidth]{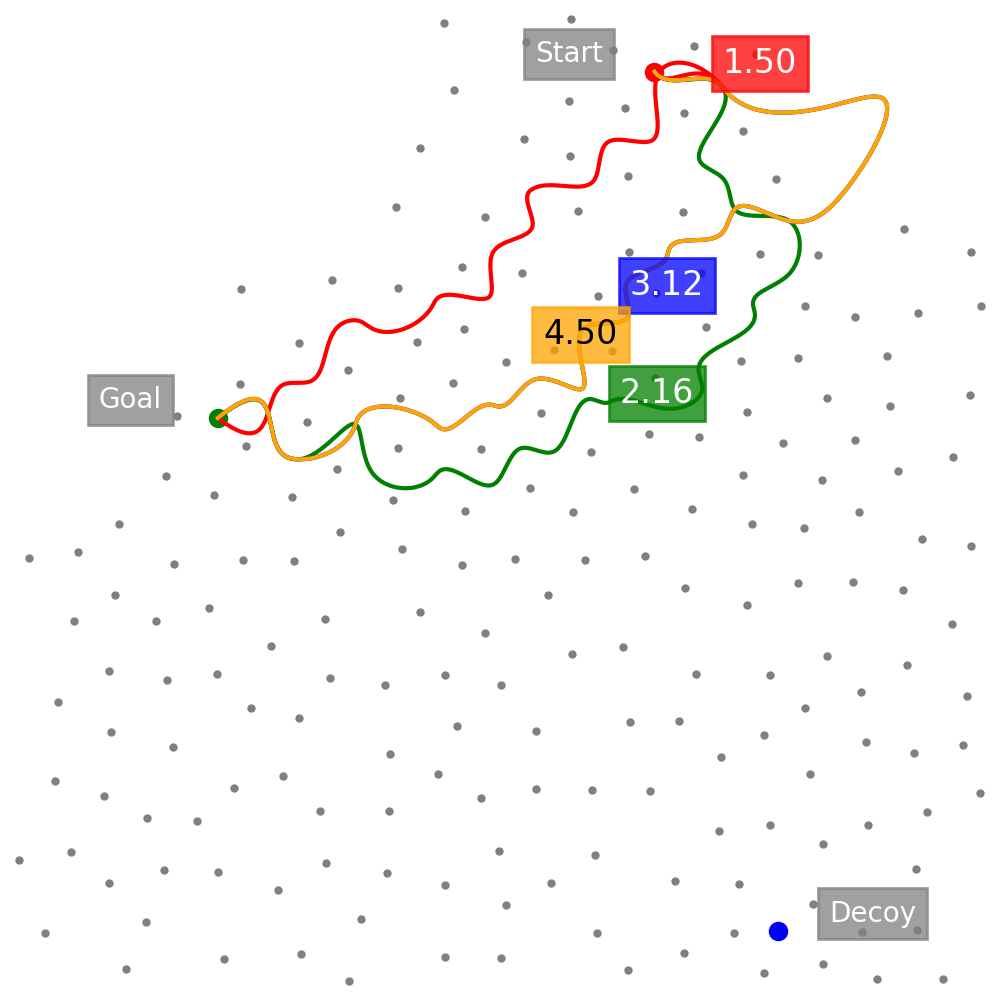}
		{Exaggeration: changing visibility (16 extra distance)}\label{fig:exag_vis}
     \end{subfigure}
     \hfill
     \begin{subfigure}[t]{0.23\textwidth}
        \centering
		\includegraphics[width=\linewidth]{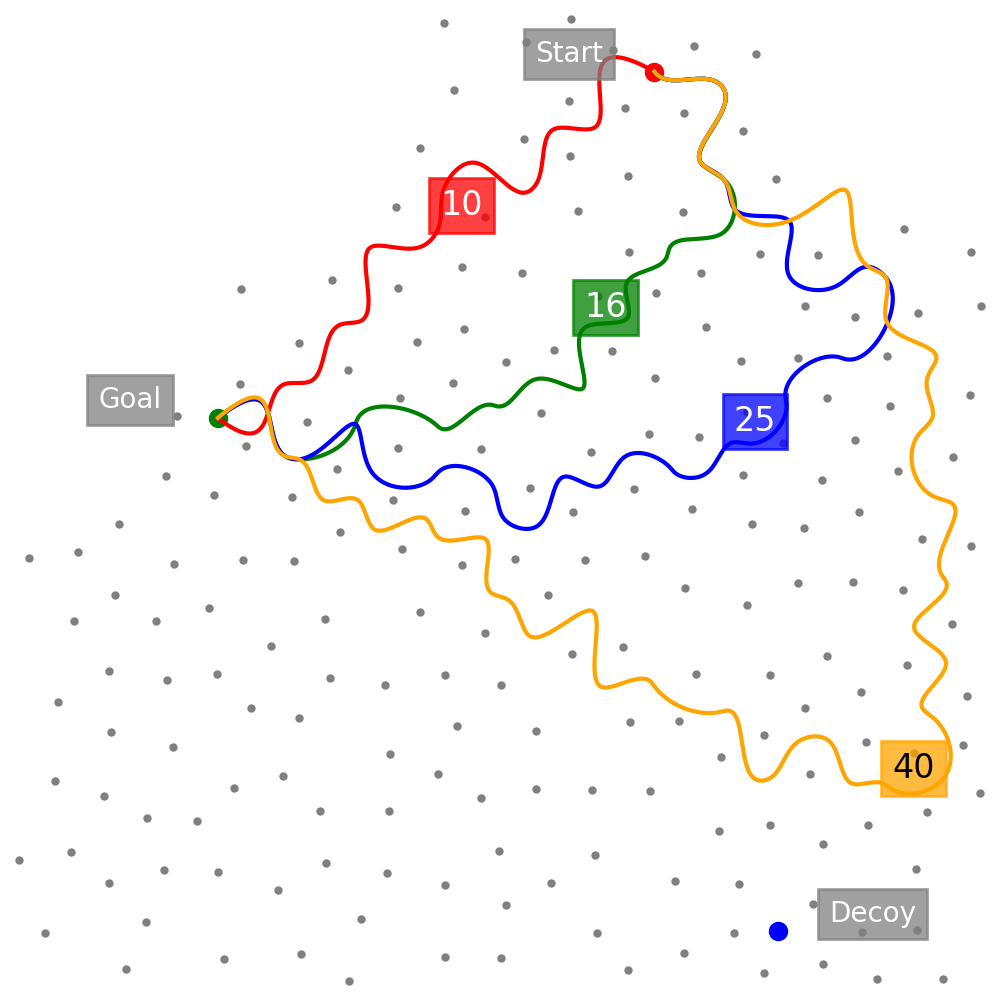}
		{Exaggeration: changing extra distance (2 visibility)}\label{fig:exag_dist}
     \end{subfigure}
     \hfill
     \begin{subfigure}[t]{0.23\textwidth}
        \centering
		\includegraphics[width=\linewidth]{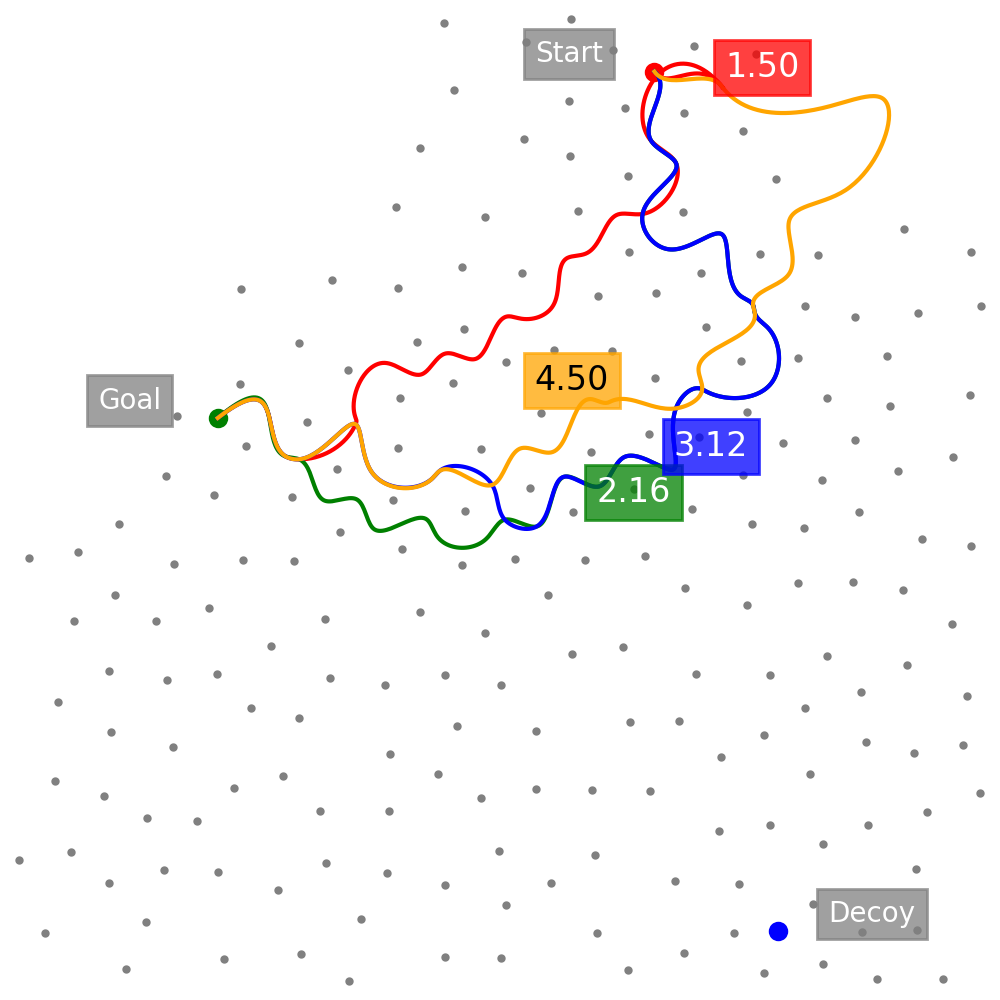}
		{Ambiguity: changing visibility (16 extra distance)}\label{fig:amb_vis}
     \end{subfigure}
     \begin{subfigure}[t]{0.23\textwidth}
        \centering
		\includegraphics[width=\linewidth]{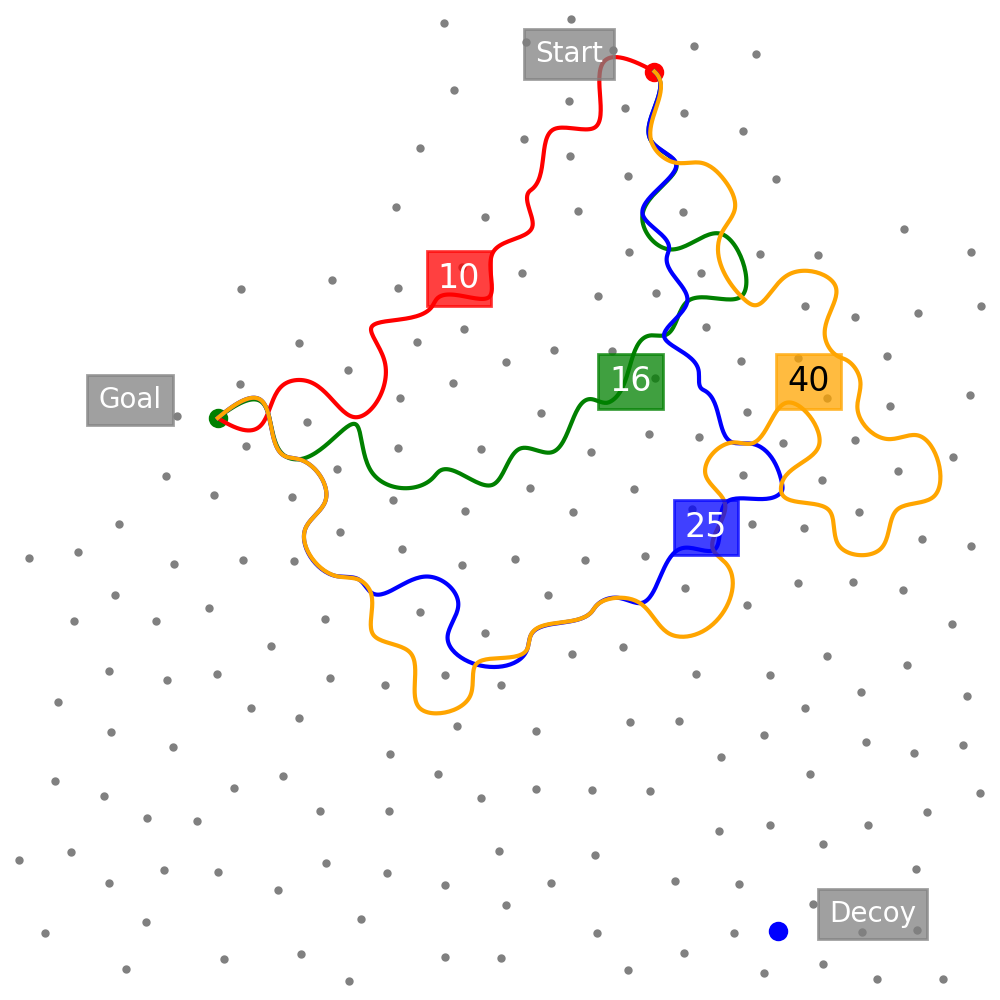}
		{Ambiguity: changing extra distance (2 visibility)}\label{fig:amb_dist}
     \end{subfigure}
        \caption{Deception tunability. When policies are given access to a wider range of nodes, their ability to act deceptively increases. For example, our results suggest that having too small of a planning radius causes deceptive capability to suffer. Meanwhile, the policy exhibits strong tunability towards path length, as paths with less urgency (more time remaining) become increasingly exaggerated, or in the case of ambiguity, add extra ``wrinkles'', perhaps as an artifact of RL training to maximize reward.}
        \label{fig:continuous_tunability}
\end{figure*}

For the results depicted in Figures \ref{fig:exag_n10_urgency} and \ref{fig:continuous_tunability}, we experimented with varying the radius of perception ("visibility") and the extra ``time-to-deceive'' that the policy is permitted to use to be deceptive.
In Figure \ref{fig:exag_n10_urgency}, we depict the difference in performance for the exaggeration policy. We chose extra levels of distance, provided to the model as a bias towards the distance remaining input parameter. We decreased the distance remaining input to the model by the Euclidean distance between consecutive points in the model's path. Because the agent operated over a discrete path, we smoothed the plots using a spline interpolation to make it easier to visualize variation between different paths. We note that because the node coordinates are not fixed to a specific grid, the graph and agent position could be resampled at arbitrary temporal resolution, allowing for quick adaptation to dynamic environments. Despite the absence of fine-tuning, we find well-correlated alignment between the human-specified path length and the true distance traveled by the agent.
In Figure \ref{fig:continuous_tunability}, we found that increasing the visibility in the continuous setting causes a positive performance increase of deception, despite the model having the same number of layers (and therefore expected modeling capacity). Additionally, increasing the allotted path length caused the model to be less conservative about the amount of exaggeration it used, and that this effect seemed to plateau for ambiguity (instead resulting in the model adding wrinkles to the path to reach the goal at the exact time limit).

\subsubsection{Dynamic Adaptivity}

\begin{figure}
    \includegraphics[width=\linewidth]{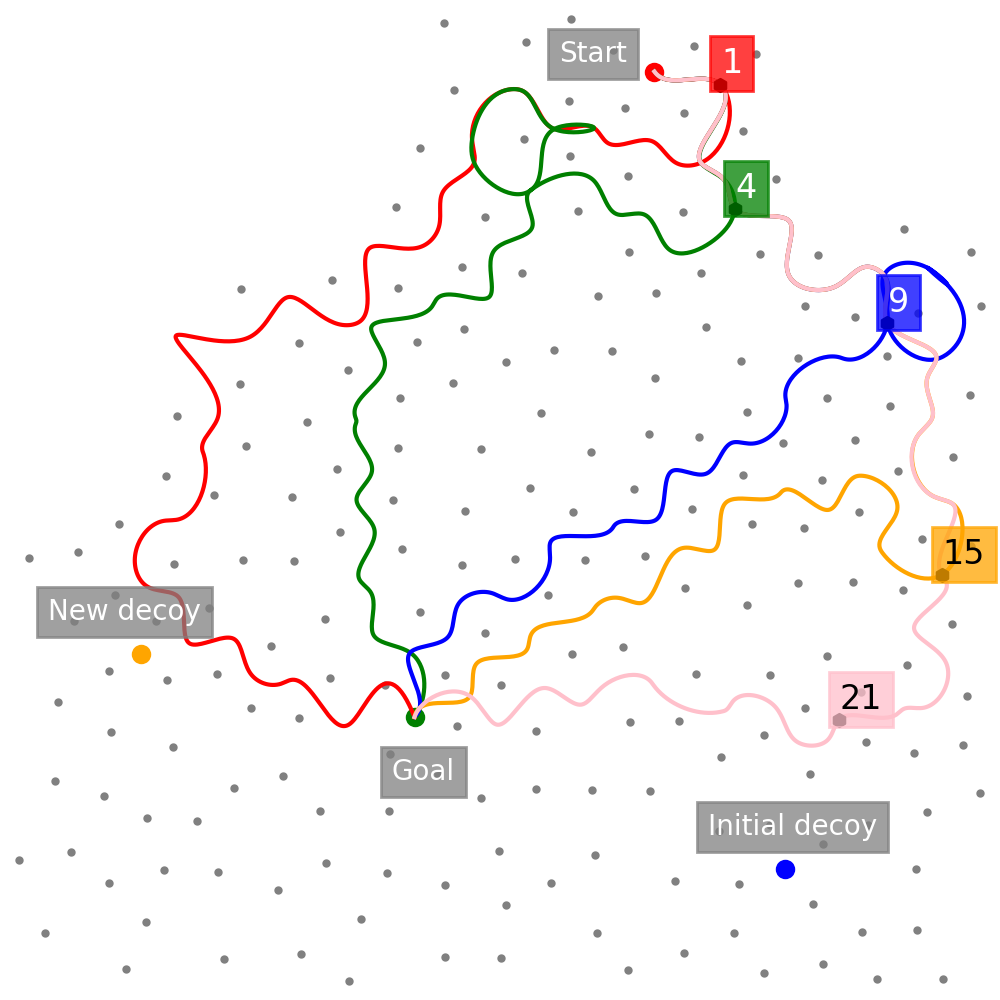}
    \caption{Dynamic adaptivity. Resulting paths from various levels of $T_{switch}$. The larger $T_{switch}$ becomes, the less exaggeration the policy applies to the new decoy. This is likely due to the policy having less time remaining to act deceptively after having consumed some extra path deviation in pursuing the original goal.}
    \label{fig:continuous_dynamic_adaptivity}
\end{figure}

To explore our exaggeration policy's ability to adapt to new goals, for the experiment depicted in Figure \ref{fig:continuous_dynamic_adaptivity} we sampled paths with $T_{max} = 30$, switching the decoy goal with a ``Plan B'' decoy goal at $t = T_{switch}$. We found that this occasionally resulted in ``S''-shaped paths, in which the agent moves towards an original decoy for the first part of the path, and then towards the Plan B decoy for the second part of the path. These results were obtained with no additional training for the policy, and using an extra distance bias of $16$.
%
%
The dynamic adaptivity just illustrated has the potential to enable other behaviors. For example, the goal and decoy positions can be changed from time step to time step, resulting in a goal that moves over time. A moving goal naturally leads to a pursuing agent, whereas a moving decoy will induce different deception into the policy. Exploring these dynamics is an interesting topic for future work.
\section{Conclusion}


In this work we have presented a novel, model-free RL scheme for training policies to perform deceptive path planning over graphs. Unlike previous methods, our approach produces policies that generalize, scale, enjoy tunable deceptiveness, and dynamically adapt to changes in the environment. We have experimentally demonstrated the advantages of our approach on a variety of gridworld problems as well as a continuous spaces navigation problem. Interesting directions for future work include extension of our method to pursuit-evasion problems over graphs and application to real-world robotic navigation tasks.




\begin{acks}
M. Y. Fatemi gratefully acknowledges the support of the University of Maryland's 2023 National Security Scholars Summer Internship Program, a cooperative agreement with the U.S. Army Research Laboratory.
\end{acks}




\bibliographystyle{ACM-Reference-Format} 
\bibliography{main}

\newpage
\appendix

\section{Appendix}




\subsection{Gridworld Environments} \label{subsec:dataset}
\begin{figure*}
    \includegraphics[width=\linewidth]{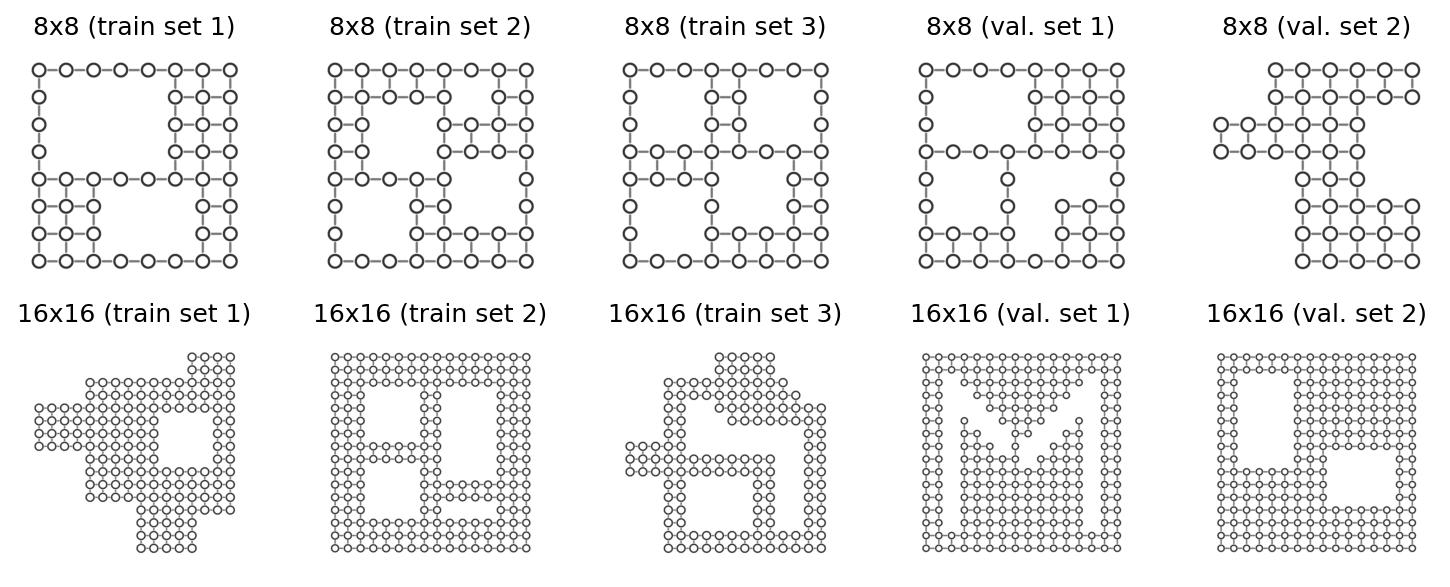}
    \caption{The set of grid worlds that were used in training and validation.}
    \label{fig:gridworlds_used}
\end{figure*}

We created the training and test gridworld environments depicted in Figure \ref{fig:gridworlds_used} using \hyperlink{Tiled}{https://www.mapeditor.org/}.

\subsection{Comparison to Linear Program Baseline}

To compare results to a linear programming baseline, we used the existing methods in \cite{savas2022deceptive} to generate paths for graphs that our neural networks had not seen in training. We used the same value function to evaluate paths from both policies. To fairly evaluate both policies, we evaluated the deterministic output from the linear program for path effectiveness and sampled $n=32$ value functions stochastically from the neural network (treating the softmax of the outputted logits for each immediate neighbor as a probability distribution for what node to move to next).

Here, we plot heatmaps of the resulting policies from the linear programming approach and graph neural network approach.

\begin{figure*}
    \minipage{0.49\textwidth}
    \centering
    \includegraphics[width=\linewidth]{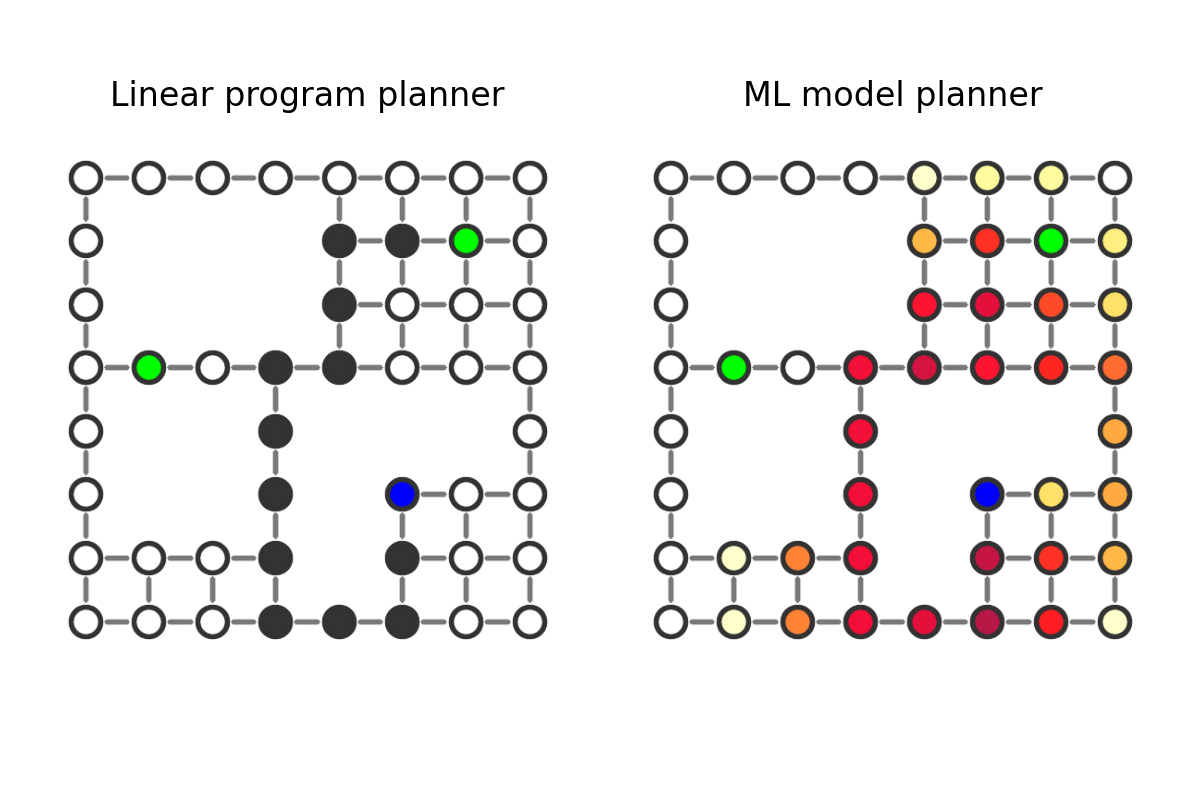}
    \caption{LP vs. proposed approach on $8 \times 8$ gridworld.}
    \endminipage
    \minipage{0.49\textwidth}
    \centering
    \includegraphics[width=\linewidth]{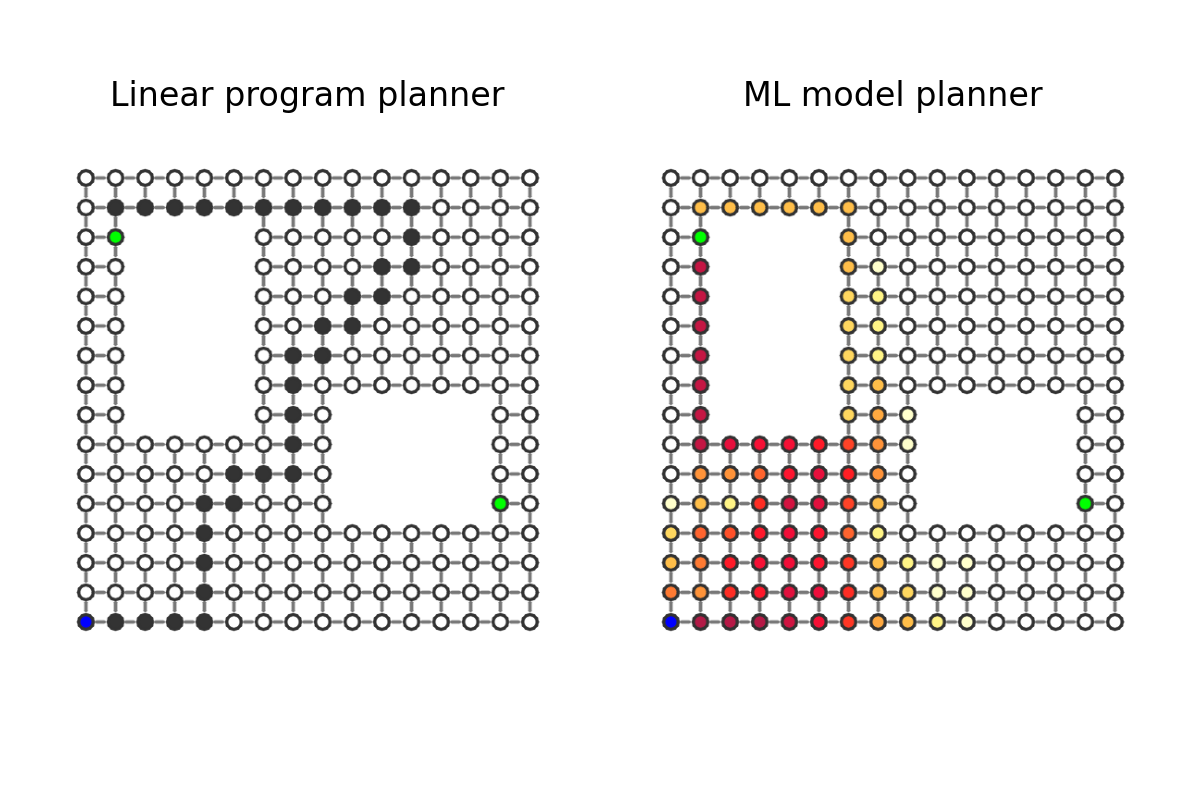}
    \caption{LP vs. proposed approach on $16 \times 16$ gridworld.}
    \endminipage
\end{figure*}

\begin{figure*}
    \minipage{0.49\textwidth}
    \centering
    \includegraphics[width=\linewidth]{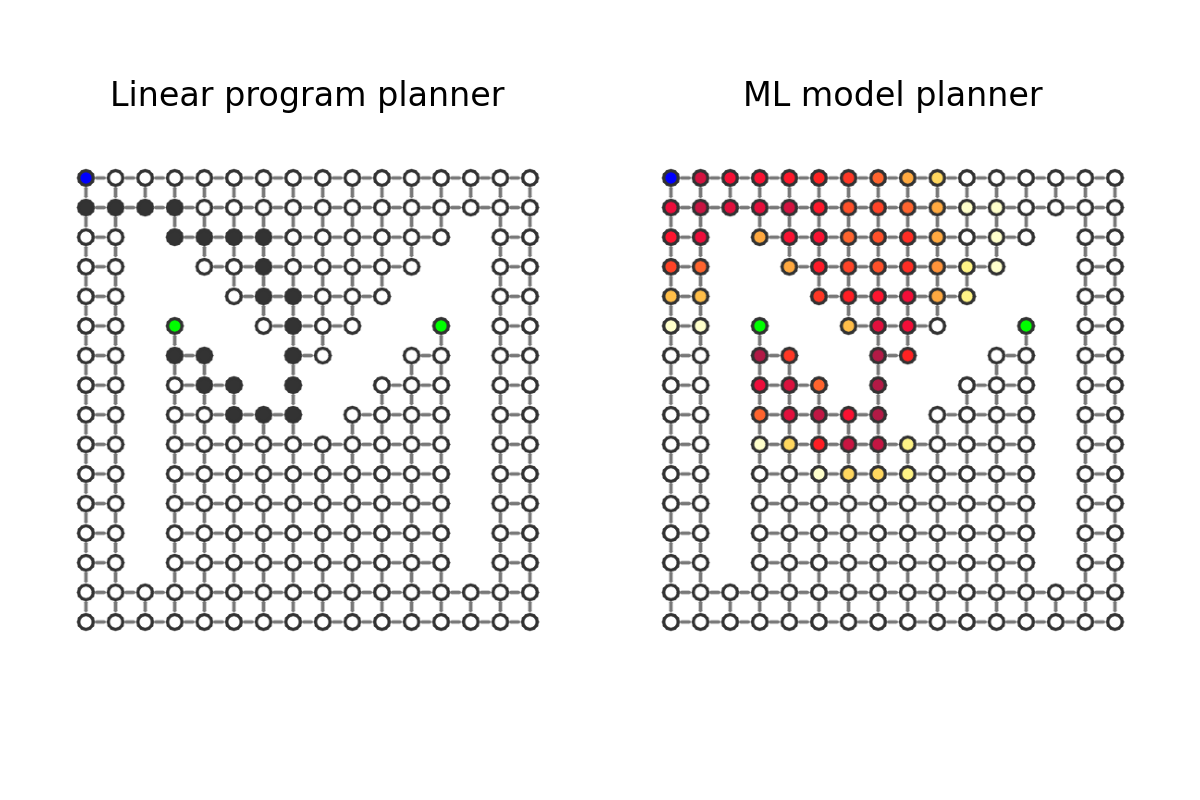}
    \caption{LP vs. proposed approach on $16 \times 16$ gridworld.}
    \endminipage
    \minipage{0.49\textwidth}
    \centering
    \includegraphics[width=\linewidth]{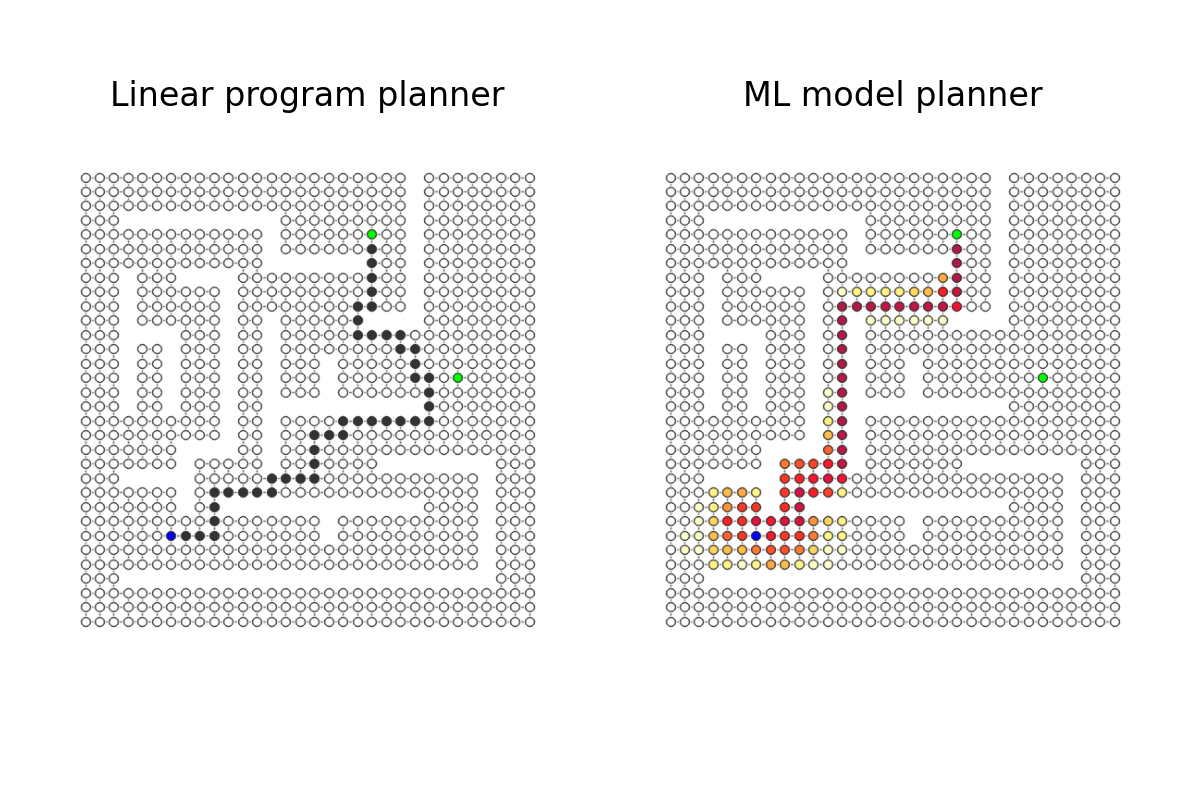}
    \caption{LP vs. proposed approach on $32 \times 32$ gridworld.}
    \endminipage
\end{figure*}


\end{document}